\documentclass[twocolumn]{bioRxiv}

\usepackage{lipsum}

\begin{document}

 \leadauthor{Ernst}

\title{Self-Supervised Learning of Color Constancy}
\shorttitle{Self-Supervised Learning of Color Constancy}

\author[ 1, \Letter]{Markus R. Ernst \orcidlink{0000-0002-2800-6346}}
\author[ 1, 2]{Francisco M. López \orcidlink {0000-0002-9653-1740}}
\author[ 1, 2]{Arthur Aubret \orcidlink {0000-0003-3495-4323}}
\author[ 3]{Roland W. Fleming \orcidlink {0000-0001-5033-5069}}
\author[ 1]{Jochen Triesch \orcidlink {0000-0001-8166-2441}}

\affil[1]{Frankfurt Institute for Advanced Studies, Frankfurt am Main, Germany}
\affil[2]{Xidian-FIAS International Joint Research Center, Frankfurt am Main, Germany}
\affil[3]{Justus Liebig University Giessen, Giessen, Germany}
\date{}

\maketitle

\begin{abstract}
Color constancy (CC) describes the ability of the visual system to perceive an object as having a relatively constant color despite changes in lighting conditions. While CC and its limitations have been carefully characterized in humans, it is still unclear how the visual system acquires this ability during development. Here, we present a first study showing that CC develops in a neural network trained in a self-supervised manner through an invariance learning objective. During learning, objects are presented under changing illuminations, while the network aims to map subsequent views of the same object onto close-by latent representations. This gives rise to representations that are largely invariant to the illumination conditions, offering a plausible example of how CC could emerge during human cognitive development via a form of self-supervised learning.
\end{abstract}


\begin{corrauthor}
mernst\at fias.uni-frankfurt.de
\end{corrauthor}

\section*{Introduction}
% Color constancy and its development
We usually perceive a banana as yellow and a stop sign as red, no matter if we see these objects under the bright midday sun or a reddish sky at dawn. This color constancy (CC) is a remarkable achievement of the visual system that takes years to develop. In fact, even 3- and 4-year-olds do not yet achieve the CC of adults \citep{rogers2020color}. Furthermore, in the same study children who had better color constancy also knew more color words. This suggests that a lack of CC may limit their ability to learn about and correctly use color words. However, what exactly are the mechanisms responsible for the development of CC?
%and how are they implemented in the brain?

% Our hypothesis of how CC is learned
While it has been shown that neural networks trained with {\em supervised} learning can replicate aspects of human color constancy \citep{flachot2022deep}, this is not a plausible model of how children acquire CC. Instead, we hypothesize that CC is at least in part acquired through a mechanism of {\em self-supervised invariance learning} as objects are seen under changing illuminations. We argue that rapid changes in the illumination of an object are abundant in our everyday life (e.g., as objects move in and out of a shadow), and that the developing visual system exploits these illumination changes to acquire CC through a self-supervised learning mechanism. This represents one example of a more general principle of exploiting the temporal structure of visual input to learn invariant representations \citep{foldiak1991learning,li2010unsupervised,wood2018development,wiskott2003slowness}.

% Sources of rapid illumination changes
Prominent examples of rapid and drastic illumination changes are clouds moving in front of the sun, an indoor light being switched on, or a curtain or door being opened or closed. While such drastic lighting changes may be few and far between, more subtle lighting changes occur much more frequently. These include a shadow falling over an object or simply a second object moving in the vicinity of the object of interest. Some of the light reflected from the second object will illuminate the object of interest and as the second object moves, this illumination will change. Since common materials reflect significant portions of the incident light (e.g.\ floors typically $>20\%$, work surfaces $20-40\%$, ceilings typically $>85\%$ \citep{steffy2002architectural}), such secondary illumination will be substantial in everyday scenes. As human adults, we are usually not aware of these effects, but they occur virtually all the time. We propose that such frequent and rapid illumination changes would in principle be sufficient to permit children to acquire CC via self-supervised invariance learning mechanisms. Here, we demonstrate the feasibility of this idea by means of a simple neural network model with temporal contrastive loss and a data set with time-varying illumination. 
% summary of contributions
We summarize our contributions as follows: 
\begin{itemize}
\setlength\itemsep{-0.5em}
    \item We propose that infants learn color constancy at least in part by exploiting the temporal structure of visual experience during changing illumination conditions.
    \item We present a new data set (C3R) consisting of colored cubes with such temporally varying illumination.
    \item We demonstrate the feasibility of our CC learning hypothesis using a contrastive learning approach with a temporal coherence loss.
    \item We evaluate the learnt representations at multiple levels in the neural network's hierarchy and compare our results to those obtained with color-based augmentations common in machine learning, where they are usually employed to make visual representations less sensitive to color.
\end{itemize}
Finally, we discuss limitations and possible extensions of the approach.

\section*{Related Work}
\subsection*{CC and its Development}
The study of color constancy in human perception dates back quite some time \citep{mollon2003origins,foster2011color}. This research has improved our understanding of the underlying perceptual and neural mechanisms. However, relatively little work has studied the {\em development} of CC during childhood and the mechanisms that drive this development. Notable exceptions are studies trying to measure CC in infants and children \citep{dannemiller1987test,rogers2020color}.

The protracted development of CC over several years \citep{rogers2020color} suggests that its establishment is a hard problem and relies on extended learning processes. Indeed, recent neural network models based on supervised learning have succeeded in reproducing certain aspects of human CC \citep{flachot2022deep,heidari2023object}. However, it is still an open question how CC could be established via more biologically plausible self-supervised learning mechanisms \citep{storrs2021unsupervised}. The development of visual processes, like CC, is not only interesting in its own right, but can also play a key role in determining the representations in the mature visual system and holds promise to be of relevance for the broader field of artificial intelligence and robotics.

\subsection*{Self-supervised Learning of Invariance}

Self-supervised learning of invariance has a long history in computational neuroscience and has recently led to major advances in computer vision, e.g. for object, action or scene recognition \citep{chen2020simple,grill2020bootstrap,feichtenhofer2021large,storrs2021learning}.
The underlying assumption of most of current approaches is that some image transformations like grayscaling significantly alter the image in the pixel space while preserving the semantic content of the image, e.g. the label and identity of an object. Crucially, these approaches rely heavily on color transformations to prevent the neural network to shortcut the learning process by extracting only the colors of the image \citep{chen2020simple}. However, none of the transformations derive from biological inspiration, nor can they account for the wide range of realistic illumination changes. This includes equalization \citep{cubuk2018autoaugment}, color channel swapping \citep{kalantari2017deep,movshovitz2016useful}, color cast \citep{movshovitz2016useful}, grayscaling \citep{chen2020simple}, solarization \citep{grill2020bootstrap} or color jittering, which modifies the brightness, contrast, saturation and hue of the image pixels \citep{chen2020simple}. Other image transformations aim to spur the extraction of shape-related features; this includes changing the apparent texture of an object or changing an image into a silhouette \citep{geirhosimagenet}. Closer to our work, some studies proposed to change the scene illumination, but without using a rendering software \citep{lo2021clcc,afifi2019else,perez2017effectiveness}. Thus, they cannot capture important illumination-related phenomena such as inter-reflections as they occur in the real world.

Our work falls within a line of research that learns invariant visual representations by exploiting the temporal structure of visual input \citep{foldiak1991learning,li2010unsupervised,wood2018development,wiskott2003slowness,fleming2019learning}. This has been used to learn object representations that are invariant to orientation \citep{schneider2021contrastive, mobahi2009deep} and positions \citep{franzius2011invariant}. In addition, this approach has been shown to support face recognition \citep{mobahi2009deep} and extracted features that better generalize over categories of objects \citep{aubreternst2023timetoaugment}. A preliminary study also showed that temporal proximity of visual inputs can support face representations invariant to illumination changes \citep{wallis2009learning}, but it is unclear whether this is directly related to the learning of color constancy. Therefore, to the best of our knowledge, we are the first to demonstrate that such an approach can support the self-supervised learning of color constancy.

\begin{figure}[htbp]
\centerline{\includegraphics[width=1\linewidth]{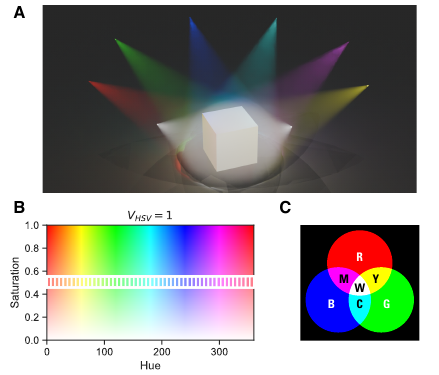}} 
\caption{Overview. A. A central cube lies on a ground plane and is illuminated by up to 8 spotlights arranged on a circle. B. Colors of the cubes in our dataset in HSV space (50 objects, $S=0.5, V=1$). C. The seven different colors (\textbf{r}ed, \textbf{g}reen, \textbf{b}lue, \textbf{c}yan, \textbf{y}ellow, \textbf{m}agenta, \textbf{w}hite) used for the spotlights.}
\label{fig:overview}
\end{figure}

\section*{Methods}
We aim to study whether self-supervised learning of time-invariant representations can elicit color constancy in the context of illumination changes. For this purpose we created a novel \emph{Color Constancy Cubes} data set with rapid temporal lighting changes (C3R, the `R' stands for ``rapid''). C3R includes 50 different objects with identical cubic shape but different reflectance properties (H,S,V values). For each object, a temporal sequence is generated (akin to a movie clip) by simulating rapid illumination changes. The changes are caused by varying the spectral composition of the light reaching the cube from different directions. 
While this dataset does not simulate the complex, richly-structured world that human infants inhabit, it provides carefully-controlled, minimal conditions for testing our main hypothesis, i.e., that it is possible to learn CC in an entirely unsupervised learning framework with passive observation.

\subsection*{Scene Setup}

% the created environment
We created different scenes using the modeling and animation software Blender (version 3.3.1). Blender can generate photo-realistic synthetic images or videos and can be conveniently controlled via an accompanying Python API. We use the BlenderProc plugin to interface with the software \citep{denninger2019blenderproc}.

Our basic setup consists of a neutral gray ground plane of dimensions $ 20 \times 20~\rm{m}^2$ upon which a cube of size $ 2 \times 2 \times 2 \, {\rm m}^3 $ resides. The cube is laid out in the middle of the plane rotated by $45^\circ$ around the z-axis. An overview of the setup is depicted in Fig.~\ref{fig:overview}A. The colors of the cubes were chosen to be of 50 equidistant hues of the HSV color space where saturation was held constant at $S=0.5$ and brightness was held at $V=1$, see Fig.~\ref{fig:overview}B. This way we ensure that the data set covers a large range of different hues while guaranteeing a certain similarity that makes it hard for the algorithm to discern the objects.

To light the scene there are eight spotlights evenly arranged in a circle  of $6~\rm{m}$ radius positioned $5~\rm{m}$ above the ground. The spotlights are directed toward the cube at a $45^\circ$ angle.
The scene camera was located at (0, -4, 4) meters and oriented at $50^\circ$ relative to the ground plane. This positioning ensures that the camera sees three sides of the cube, which may all be illuminated differently. A detailed description of the remaining rendering and object parameters is given in the \hyperref[sec:appendix]{Appendix}.

\subsection*{Generation of Sequences}

\begin{figure}[htbp]
\centerline{\includegraphics[width=1.0\linewidth]{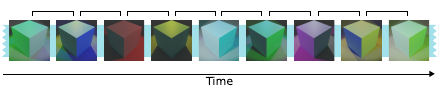}}
\caption{
The temporal structure of the C3R data set. Each of the eight spotlights is randomly assigned a color and luminosity. The resulting illumination rapidly changes over time and gives different impressions of the colored (blue) cube in a sequence. 
The algorithm shapes the latent representation of impressions in temporal vincinity (square brackets) to be alike.
}
\label{fig:temporal_dataset}
\end{figure}

The sequences for each object are generated in three steps by adapting the lighting conditions, i.e. modifying the spotlights from frame to frame, as illustrated in Fig.~\ref{fig:temporal_dataset}.  

First, spotlights only turn on with probability $p_{\rm{on}}=0.5$, otherwise they remain off for the current frame. This way, the cube on the ground will be illuminated by a different number of spotlights from different directions and is not fully illuminated from each side all the time.
Second, we simulate spectral composition changes by sampling one of 7 predefined colors to tint each spotlight, displayed in Fig.~\ref{fig:overview}C. The specific colors are chosen because their mixture covers the whole color space.
Third, for each spotlight that lights up we sample a power $P$ between $300$ and $1000$ W. We lower-bound the power by $300$ W to avoid obtaining a mostly dark scene and upper-bound it by $1000$ W considering the existence of equivalent fixtures for interior lighting. As a result, the intensity of the light reaching the cube can vary drastically.
After parameters for all 8 spotlights are chosen the process starts over for the next frame of the sequence until the required images per object are generated.
The resulting sudden lighting changes are rather extreme. We chose this setting because we suspect that such drastic illumination changes may be ideally suited for learning CC. One interpretation of the colored spotlights is that they model indirect illumination of an attended object by colored objects in the vicinity. Note, however, that such sudden and drastic illumination changes would be unlikely to occur during an infant's daily visual experience. Our goal is to establish a first proof of concept rather than faithfully modeling everyday visual experience.

The generated data set contains $50{,}000$ images split into 50 sequences (one $1{,}000$-image sequence per object). Each frame in the sequence shows the corresponding cube illuminated by a different setting of the spotlights. 
$60\%$ of each sequence, i.e., $30{,}000$ images, form the training set and  $20\%$, i.e. $10{,}000$ images each, form the validation and test set. Images are rendered at $32 \times 32$ pixels \footnote{Data sets and code are made available online at \\ \href{https://github.com/trieschlab/ColorConstancyLearning}{https://github.com/trieschlab/ColorConstancyLearning}}.

\subsection*{Neural Network Training and Evaluation}

\begin{figure}[htbp]
\centerline{\includegraphics[width=0.95\linewidth]{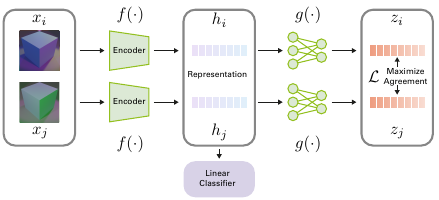}}
\caption{The self-supervised learning framework used for this study. Two successive images are encoded by a neural network $f(\cdot)$ to yield a hidden representation $h$ that is used for downstream classification tasks. The network is trained by projecting the hidden representation using a projection head $g(\cdot)$ and maximizing agreement in the resulting space $z$.}
\label{fig:sslframework}
\end{figure}

\paragraph{Self-supervised training}
To learn image representations, we use a previously introduced time-contrastive learning method, SimCLR-TT \citep{schneider2021contrastive}, summarized in Fig.~\ref{fig:sslframework}. At each learning iteration, SimCLR-TT samples a batch $\mathcal{X}$ containing $N$ images and their $N$ immediate successors in time, resulting in a batch of size $2N$. 
Each pair $({x}_i, {x}_j)$ in the batch is passed through the neural network $f(\cdot)$ and the projection head $g(\cdot)$. Afterwards, the embeddings are updated via self-supervised learning. Specifically, the image embeddings $f({x}_n)$ with $n\in i,j$ are updated by changing the weights of the neural network $f(\cdot)$ and the projection head $g(\cdot)$ to symmetrically maximize the agreement of the projections of an image and its successor $z_{n} = g(f({x}_{n}))$  with $n\in i,j$ via back-propagation:
\begin{equation}
    \mathcal{L}(z_i,z_j)= -\log \left( \frac{  e^{ \left( {\rm cos}(z_i,z_j)/\tau \right)}  }{\sum_{k=1,k \neq i}^{2N}  e^{\left({\rm cos}(z_i,z_k)/\tau \right)} } \, \right) ,
    \label{eq:simclr}
\end{equation}
where $\tau$ is a temperature hyper-parameter and ${\rm cos}$ is the cosine similarity with high value indicating high similarity. For our experiments we choose $\tau = 1$. The numerator in the fraction in Eq. \ref{eq:simclr} promotes that successive embeddings share similar representations while the denominator prevents a collapse of the embeddings into a single point in the representation space by imposing that all representations are dissimilar. 

In practice, we implement $f(\cdot)$ as a convolutional neural network (CNN) similar to LeNet5 \citep{lecun1989backpropagation} and adapt it for self-supervised learning. The encoder $f(\cdot)$ has two convolution layers $l_1$ (6, 5, 1, 0) and $l_2$ (16, 5, 1, 0), where the parameters stand for (output channels, kernel size, stride, padding). After each convolution layer, we apply a ReLU and a MaxPool2D operation. The two convolutional layers are followed by two linear layers $l_3$, $l_4$ with 120 and 84 neurons, respectively. These layers are again interspersed with ReLU activation functions and the output of $l_4$ constitutes the learnt latent representation $h$. As commonly done in self-supervised learning \citep{balestriero2023cookbook}, we add a fully connected projection head $g(\cdot)$, with one hidden layer of 128 neurons, on top of the CNN backbone. 
For network training we choose the following hyper-parameters: a learning rate of $0.001$, AdaM optimizer and a batch-size of $2N=600$. We train each encoder model for 100 epochs. Computation was done on GPUs of type NVIDIA GeForce RTX 2070 Super and RTX 2080 Ti.

\begin{figure*}[htbp]
\centering
\includegraphics[width=1.0\linewidth]{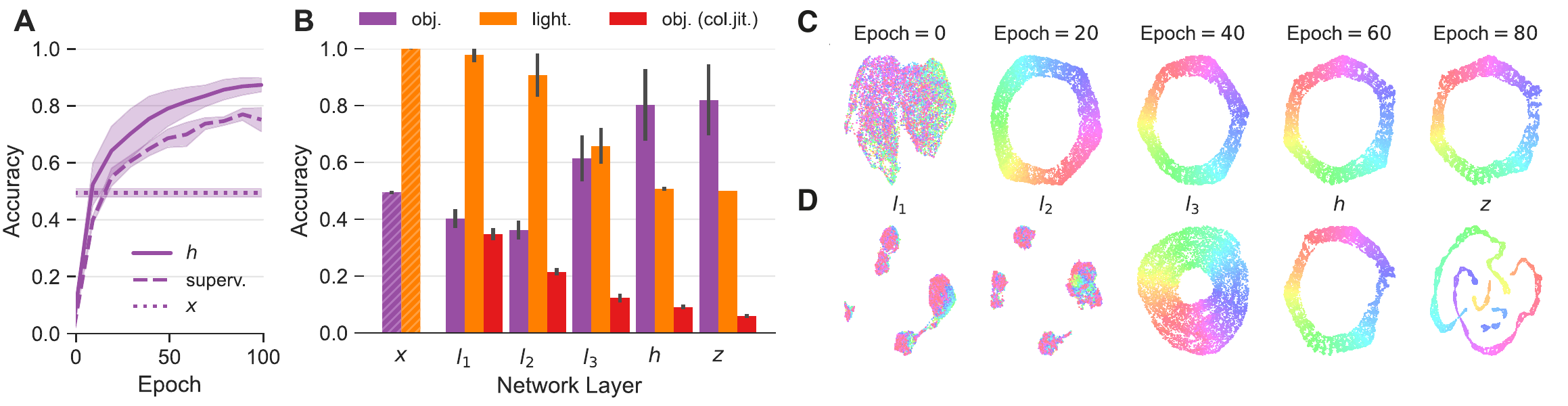}\\

\caption{Overall results. A. Learning curve of our proposed method (solid) compared to a supervised baseline (dashed). Horizontal line (dotted) shows raw pixel accuracy. B. Downstream classification of object colors and lighting at different layers of the network hierarchy for our approach vs. object accuracy from pure color jittering (red). C. PaCMAP visualization of the representation $h$ evolving during training. D. Representation at different points of the network hierarchy after training has completed (epoch = 100). Error bars and envelopes in A and B depict standard deviation from five independent runs.}
\label{fig:results_overall}
\end{figure*}

\paragraph{Evaluation}

To evaluate the learnt representation, we use the standard protocol of self-supervised learning \citep{chen2020simple}. Thus, after self-supervised training, we freeze the weights of the CNN, extract the CNN features (before the projection head by default) and train a linear classifier in a supervised fashion to predict the cube identities. The classifier aims to assess whether the latent representation captures color constancy, i.e., an embedding where frames from the same object can be clearly separated from objects with different colors. When classifying cube identities we represent the target vector as a one-hot vector with the corresponding class-index being a one and everything else being zero and use a cross-entropy loss. When using the representation to classify lighting conditions we represent lighting as a 24-dimensional n-hot vector where each triplet of indices corresponds to a binary RGB value of each light, e.g. magenta is $(1,1,0)$. Note that this discards information regarding the precise powers of the light sources (between 300 and 1000 W). To cope with multi-labels we employ a binary cross-entropy loss and keep all remaining parameters fixed. Every linear readout is trained for 200 epochs.

To obtain a simple baseline comparison to our results we also investigated learning CC solely from color augmentations. Here we start out with a single equally lit (all spotlights on, 500 W, white light) render of all 50 objects as a base. We then apply a random color jitter transformation commonly used in self-supervised learning. This transformation generates two augmented images $(x_i,x_j)$ from the same source but with different color properties. In our case we apply color jittering changes in the range $(0.8,0.8,0.8,0.2)$ for (brightness, contrast, saturation, hue) as described in \citep{chen2020simple} and then proceed with ordinary SimCLR learning.

Every training and evaluation procedure was repeated five times with different random seeds.
To statistically compare different networks and their performance we used a two-sided Student's T-test \citep{student1908ttest} with a Bonferroni correction for multiple testing ($\alpha = .05$).

\section*{Results}
In this section, we present the results obtained with the models.
Figure~\ref{fig:results_overall} depicts the overall results of our experiments. The learning curves in Fig.~\ref{fig:results_overall}A illustrate that learning color constancy with our self-supervised temporal objective succeeds and surpasses a linear evaluation of the raw pixel representation (dotted horizontal line). In addition, our technique also clearly outperforms a purely supervised baseline (dashed) trained with exactly the same parameters, but with a cross-entropy loss function.

To gain more insight into how the representation evolves within the network hierarchy, we probed the different layers of the network with a linear readout. The results of this analysis can be seen in Fig.~\ref{fig:results_overall}B. In addition to classifying the object color (purple) we also classified lighting conditions (orange). 
For object color classification (our test of color constancy) the representation in the lower layers $l_1$, $l_2$ give poor results. Classification rates remain below those of the baseline obtained by classifying based on the raw pixel values ${x}$. In contrast, the cardinal representation $h$ permits classification significantly outperforming the baseline ($t(8) = 4.85$, $p = .001$). Classification accuracies based on the learnt representation $h$ vs.\ the projection $z$ that we directly optimize are comparable ($t(8) = -0.20$, $p = .84$).

As for lighting classification we almost observe a mirrored image. Here, the lower layers $l_1$, $l_2$ retain the most information regarding the lighting which seems to get lost from layer to layer. In fact, we observe that the classification accuracy for $h$ is significantly lower than classifying from raw pixels ${x}$ ($t(8) = -156.14$, $p < .001$). Nevertheless, the accuracy never drops below 50\%, whereas randomly guessing the color of the spotlights would yield an accuracy of $0.125$.
Additionally, we trained a representation purely based on the color jitter augmentation (red) and evaluated it the same way. Here only object color classification is possible since the lighting is held constant.
Color jittering was not able to capture the intricacies of CC in a suitable representation, but still performs considerably better than random guessing ($2\%$). When comparing to our approach, the C3R-based learning at $h$ significantly outperforms color jittering at the same network level ($t(8) = 11.23$, $p < .001$).

In Fig.~\ref{fig:results_overall}C,D we applied a PaCMAP dimensionality reduction \citep{wang2020pacmap} on the learnt representation to obtain an additional qualitative view. Each point corresponds to a particular object seen under a particular illumination from the test set. The colors of the points correspond to the true colors of the objects as shown in Fig.~\ref{fig:overview}B. We observe that relatively early on in training (Epoch 20) the representation $h$ converges to a circular pattern representing the gradual shifts in hue of the object color. Over time the representation additionally shows signs of granular clustering making it easier to discern the individual classes in a downstream task. Clusters corresponding to objects of similar color occasionally show some overlap. This reveals that the network successfully learned an orderly clustering of images according to the intrinsic color of the depicted object, despite substantial changes in pixel values caused by the illumination. Thus, the representation reflects color constancy.

Figure~\ref{fig:results_overall}D depicts the representations within the network hierarchy after training has been completed. While the first layers lack a clearly defined structure, their output representation already shows ``pockets'' of classes that cluster together. Layer $l_3$ starts to distribute classes in a meaningful, circular way and the latent representation $h$ again shows the now fully formed pattern with granular clusters. The representation of the projection layer $z$ reveals a chain-like structure that coagulates some but not all of the 50 classes.

\begin{figure}[htbp]
\centerline{\includegraphics[width=1.0\linewidth]{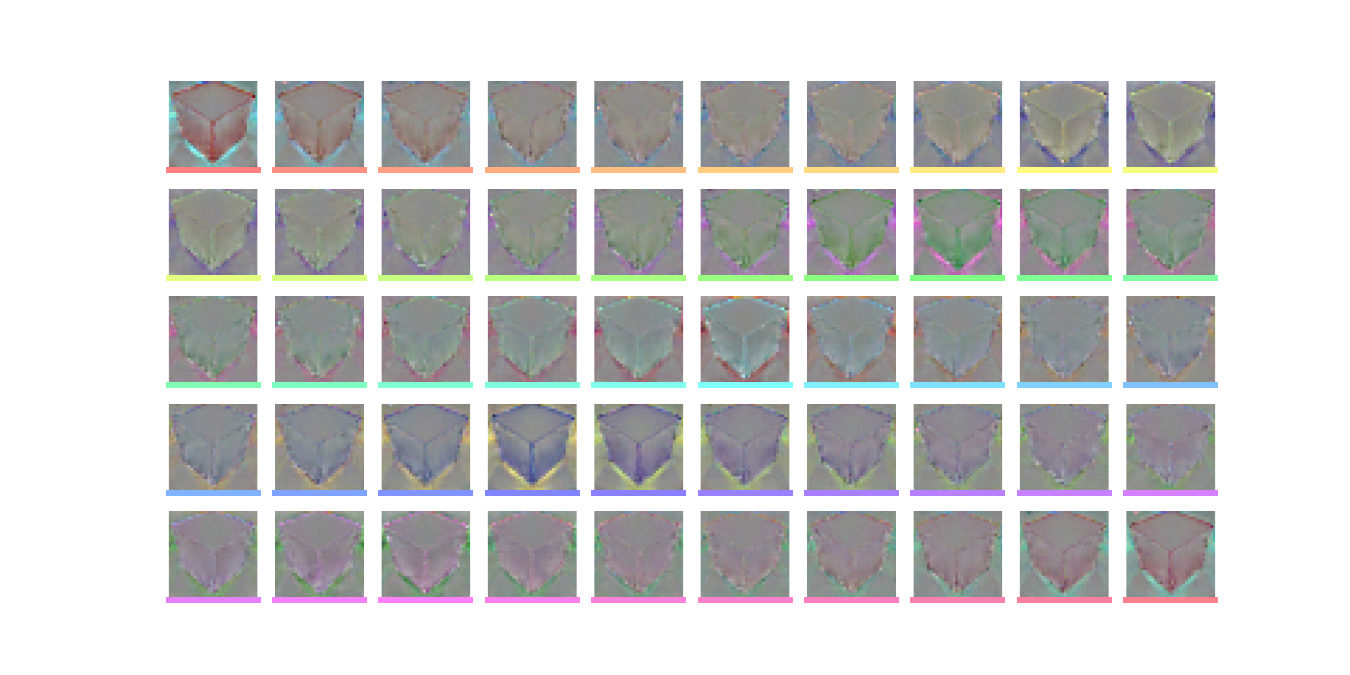}}
\caption{Linear weights visualized in RGB space for raw pixel classification after 200 epochs of training. Brighter means higher value. Colored underline depicts the true color of each object.}
\label{fig:linear_weights}
\end{figure}

It is interesting that the representation learned via SimCLR-TT retains some lighting information, although the network aims to become invariant to lighting variation. One reason for this may be the ground plane, which provides a neutral reflective surface from which the lighting can be inferred relatively easily. In fact, in order to achieve CC, the lighting information must be incorporated as context information to infer the true object color/class. The special role of the ground plane as providing context to infer the lighting situation is also revealed by considering the neural network weights for object classification from raw pixels shown in Fig.~\ref{fig:linear_weights}. In this visualization, each weight that maps to a certain class contributes to their respective color channel. I.e., the three weights from an RGB pixel to the output node for a particular class are represented as a single RGB pixel whose R, G and B values correspond to these three weights. We observe that the overall coloring of the cube is important, since the weights reflect almost the same hue as the true object color.
%outline.
Furthermore, the area where the cube touches the floor contains oriented color opponent structures, e.g., blue-yellow or red-green transitions. This can be understood in the following way: when the light reflected from the lower border of the cube is, say, more red than that reflected from the ground plane right next to the cube, this provides evidence for the cube being red. The importance of this mechanism is reflected in the rather strong weights (visible as highly saturated colors) where the cube touches the ground plane.

\section*{Discussion}
% What was our contribution?
We have proposed that color constancy emerges during human development at least in part through a form of self-supervised temporal invariance learning. Observing objects under varying illuminations permits, in principle, to extract what remains invariant, i.e., the reflectance properties of the object, and disentangling it from the more rapidly changing illumination parameters. We have presented a simple model based on temporal contrastive learning to demonstrate the feasibility of the theory. Previous work had shown that neural network models can replicate aspects of human CC \citep{flachot2022deep,heidari2023object}, but these models have relied on supervised learning approaches, which are implausible as models of the {\em acquisition} of CC during human development. 
Here we sought to test under highly-restricted conditions whether unsupervised contrastive learning in time is in principle sufficient for to acquire illumination-invariant color representations.

Our experiments showed that the proposed approach learns representations that reflect meaningful color properties and provide at least comparable results to a supervised learning system. 
As to why our approach worked best, we are unsure. For a start it could be a matter of choosing the right hyperparameters, however it could also point to a larger issue where the way we encode class labels as diametrically opposed contradicts the internal structure of the data, where neighboring hues are close by. % than e.g. blue and red.  
We observed that object instance information is slowly built up from layer to layer. This can be explained by the convolutional nature of the used network. As units in lower layers only have limited receptive field sizes combining global information is not possible. At $h$ the representation is consolidated again after two linear layers recovered the global structure.
Inferring and exploiting lighting information in the network’s representation is crucial to successfully master the given task. This is in line with our finding that the color jittering experiments gave relatively poor results. 

When considering direct classification from the raw pixel images and examining the weights that map the pixel space to the different classes (Fig.~\ref{fig:linear_weights}), we observed color opponent structures at the interface of cube and ground plane. This is because the contrast between the cube and the floor provides critical evidence to determine the true color of the shown object. For instance, irrespective of the drastic illumination changes, the lower border of a blue cube is more bluish than the gray ground it resides on and similar for all other queried colors.

% Limitations of the current model
Our model has a number of limitations. Most importantly, the visual environment in which the system learns is greatly simplified compared to the visual experience of real infants. Our model's learning environment consists of objects of identical (cubic) shape seen always from the same perspective. Furthermore, the lighting changes experienced by the system  (switching colored spotlights) are more extreme and varied compared to lighting changes one would expect to observe during children’s everyday interactions with objects. In this sense, our setup still represents a highly idealized situation that offers a proof-of-concept rather than being a faithful model of the development of CC in humans. It would be interesting to test conditions in which the temporal (and wavelength) statistics more closely resemble the human environment, e.g., with a mixture of diurnal cycles and higher frequency changes.

The model reaches an impressive level of color constancy that may be well beyond human abilities. Considering the example scenes in Fig.~\ref{fig:temporal_dataset}, we speculate that human subjects may be substantially worse at identifying the correct object among the set of 50 objects based on a single image. We suspect that the reason for this are the various idealizations mentioned above.

A more realistic model should consider lighting, shape, reflectance and viewpoint changes that are modeled more closely after those occurring during natural interactions with objects. Typically, humans do not even notice these more subtle illumination changes. We speculate that the CC mechanisms operating in adults may in fact be the chief reason why they do not notice such subtle illumination changes (anymore). We also speculate that the subtlety of most everyday illumination changes contributes to the rather slow development of color constancy taking several years \citep{rogers2020color}. Next to modeling typical lighting changes more accurately, it could also be interesting to introduce more diverse material properties to the objects (different amounts of specularity, sub-surface scattering, etc.).

% Discuss from a broader perspective
The particular self-supervised learning method we have used is a form of temporal self-supervised learning \citep{schneider2021contrastive}. Interestingly, this approach has also yielded promising results in models of the development of human object recognition abilities \citep{schneider2021contrastive, aubreternst2023timetoaugment}. Therefore, the joint learning of object representations and CC is an interesting direction for future work. The easiest way to achieve this might be to incorporate additional object shapes and show the objects from different perspectives. We expect such a system to learn a joint encoding of object shape, pose, and color and exhibit different degrees of invariance to these factors depending on how frequently these properties are changing.

\section*{Acknowledgments}
This research was funded by the Hessian Ministry for Higher Education, Research, Science and the Arts (Cluster Project ``The Adaptive Mind''), the Deutsche Forschungsgemeinschaft (DFG project ``Abstract REpresentations in Neural Architectures (ARENA)''), as well as the European Research Council (ERC-AdG-2022, ``STUFF'': 101098225). JT is supported by the Johanna Quandt Foundation.
We thank Ali Can Kara for initial contributions to this research. 

\section*{Bibliography}	
\bibliography{references.bib}
\bibliographystyle{IEEEtran.bst}

%\newpage

\appendix
\section*{Appendix}
\label{sec:appendix}
Here we describe additional settings of our Blender rendering pipeline.
For rendering we use Blender's physically-based path tracer {\em Cycles}. Rendering of the images is based on 64 samples and the built-in denoising algorithm.
The material properties of the cubes and the floor are identical and defined by their bidirectional scattering distribution function (BSDF). We use the Blender (3.3.1) built-in {\em principled BSDF} node, which combines many shading options and provides sensible defaults. The so-called {\em alpha} parameter sets the transparency/opacity of an object.  We choose the value 1, which corresponds to a completely opaque object.
{\em Sheen} creates soft, velvet-like reflections near the edges. Values range from 0 to 1 and we use a value of 0.
The {\em sheen tint} determines to what extent the edge reflections have the color of the light source ($\text{sheen tint} = 0$) or of the object ($ \text{sheen tint} = 1$). We chose an intermediate value of 0.5.
{\em Roughness} indicates the micro surface roughness and was chosen as 0.5. Additionally we chose 
%{\em sheen} = 0, {\em sheen tint} = 0.5, 
$\text{{\em clearcoat roughness}} = 0.03$, and the Index of Refraction ($\text{\em{IOR}} = 1.45$). {\em Specular reflection}, {\em metallic reflection}, {\em subsurface scattering}, {\em clear coat} as well as {\em light emission} were not used. The result is a medium roughness and homogeneous material with no sub-surface scattering.

%cube_mat.node_tree.nodes['Principled BSDF'].inputs['Base Color'].default_value = (1, 1, 1, 1)
%cube_mat.node_tree.nodes['Principled BSDF'].inputs['Specular'].default_value = 0.5
%cube_mat.node_tree.nodes['Principled BSDF'].inputs['Roughness'].default_value = 0.5
%cube_mat.node_tree.nodes['Principled BSDF'].inputs['Sheen Tint'].default_value = 0.5
%cube_mat.node_tree.nodes['Principled BSDF'].inputs['Sheen'].default_value = 0.0
%cube_mat.node_tree.nodes['Principled BSDF'].inputs['Clearcoat'].default_value = 0.0
%cube_mat.node_tree.nodes['Principled BSDF'].inputs['Clearcoat Roughness'].default_value = 0.03
%cube_mat.node_tree.nodes['Principled BSDF'].inputs['Metallic'].default_value = 0.0
%cube_mat.node_tree.nodes['Principled BSDF'].inputs['Transmission'].default_value = 0.0
%cube_mat.node_tree.nodes['Principled BSDF'].inputs['Transmission Roughness'].default_value = 0.0
%cube_mat.node_tree.nodes['Principled BSDF'].inputs['IOR'].default_value = 1.45

%\include{02_Article_Supplementary}
%\include{03_Article_SupplementaryVideos}

\end{document}